\documentclass{article}

\usepackage{arxiv}

\usepackage[utf8]{inputenc} 
\usepackage[T1]{fontenc}    
\usepackage{hyperref}       
\usepackage{url}            
\usepackage{booktabs}       
\usepackage{amsfonts}       
\usepackage{nicefrac}       
\usepackage{microtype}      
\usepackage{graphicx}
\usepackage[numbers]{natbib}
\usepackage{doi}

\graphicspath{ {./figs/} }
\usepackage{algorithm}
\usepackage{algpseudocode}
\usepackage{wrapfig}
\usepackage{caption,subcaption}
\usepackage{array}
\usepackage{amsmath}

\title{Low Resource Summarization using Pre-trained Language Models}

\date{}	

\author{ \href{https://orcid.org/0009-0008-2938-3590}{Mubashir Munaf} \\
	\texttt{mubashir.munaaf@gmail.com} \\
	\And
	\href{https://orcid.org/0000-0001-9583-5585}{Dr Hammad Afzal} \\
	\texttt{hammad.afzal@mcs.edu.pk} \\
    \And
	{Naima Iltaf} \\
	\texttt{naima@mcs.edu.pk} \\
    \And
    {Khawir Mahmood} \\
	\texttt{khawir@mcs.edu.pk} \\
}

\renewcommand{\shorttitle}{Low Resource Summarization}

\hypersetup{
pdftitle={Low Resource Summarization},
pdfsubject={NLP},
pdfauthor={Mubashir, Hammad et al.},
pdfkeywords={Automatic Summarization, Low Resource Language, Low Resource Summarization, Multilingual Models, Urdu Summarization, Natural Language Generation (NLG), Natural Language Processing (NLP)},
}

\begin{document}
\maketitle

\begin{abstract}
With the advent of Deep Learning based Artificial Neural Networks models, Natural Language Processing (NLP) has witnessed significant improvements in textual data processing in terms of its efficiency and accuracy. However, the research is mostly restricted to high-resource languages such as English and low-resource languages still suffer from a lack of available resources in terms of training datasets as well as models with even baseline evaluation results. Considering the limited availability of resources for low-resource languages, we propose a methodology for adapting self-attentive transformer-based architecture models (mBERT, mT5) for low-resource summarization, supplemented by the construction of a new baseline dataset (76.5k article, summary pairs) in a low-resource language Urdu. Choosing news (a publicly available source) as the application domain has the potential to make the proposed methodology useful for reproducing in other languages with limited resources. Our adapted summarization model \textit{urT5} with up to 44.78\% reduction in size as compared to \textit{mT5} can capture contextual information of low resource language effectively with evaluation score (up to 46.35 ROUGE-1, 77 BERTScore) at par with state-of-the-art models in high resource language English \textit{(PEGASUS: 47.21, BART: 45.14 on XSUM Dataset)}. The proposed method provided a baseline approach towards extractive as well as abstractive summarization with competitive evaluation results in a limited resource setup.
\end{abstract}


\section{Introduction}
\label{sec:intro}
Natural Language Processing (NLP) is a sub-field of linguistics \& computer science with an addition of artificial intelligence. Mostly comprises of two distinct portions of \textit{Natural Language Understanding (NLU)}; understanding and extracting meaningful insights of natural languages and Natural Language Generation (NLG) \cite{Gatt2018} generating content similar to human language for desired tasks. Most of the research was initially focused on NLU however in the recent past after the inception of advanced architectures, generation methods have also improved. Automatic Summarization can be broadly categorized into Extractive and Abstractive Summarization \cite{Aggarwal2015}. In \textit{Extractive Summarization} summary of the content is generated by selecting or extracting part of the content from the original input content. This content can be sentences, paragraphs, or part of sentences/phrases. In \textit{Abstractive Summarization} summary is generated usually by NLG techniques. An abstractive summary of an input content/document comprises new words, phrases, or sentences which are not present in the input document and are generated based on the meanings or context of an input document. \textit{Hybrid or Aided Summarization} is also used in which a combination of extractive and abstractive approaches or Human aid is utilized in the form of topics, phrases, or bias.

Early research comprised statistical methods for summarization involving a bag-of-words approach (considering words individually without order in a sequence) \cite{Luhn1958}\cite{Jones1972}\cite{Edmundson1969} lacking contextual information. With the inception of Deep Learning (Artificial Neural Networks; ANN) based models, word orders and their dependencies in a sequence \cite{Rush2015}\cite{Bahdanau2014}\cite{Sutskever2014}\cite{Nallapati2016}\cite{See2017} usually sentences and paragraphs were addressed. However, these models involved excessive training with requisite datasets (document/summary pairs) and computational resources (GPUs) which are often rarely available unless in a specialized established environment.

Pre-trained Language Models based on ANN utilizing transfer learning \cite{Devlin2019}\cite{Radford2018}\cite{Raffel2020} gave the opportunity to re-use a Language Model (LM) trained with excessive generalized training data \textit{(self-supervised training; without the need of explicit document/summary pairs)} capable of understanding natural language and its representation. These learned representations of language can be implemented with less training or even zero shot settings to various downstream tasks e.g. Summarization. Most of the research is focused on high-resource languages like English. Lack of research in low-resource languages make it difficult to develop NLP models when even baseline datasets, models and evaluation results are not available. Recently multi-lingual models have been trained \cite{Devlin2019}\cite{Artetxe2018}\cite{Lample2019} to address such impediments. Although multilingual models suffer from the disadvantage of under-representation of low-resource languages \cite{Pires2019} still provide an opportunity to explore such avenues to produce competitive baseline research for low-resource languages.

\subsection{Research Objectives}
Our research objectives are deliberated towards an area with limited exploration which is lacking sufficient NLP resources (datasets and models). Therefore, we aimed at producing competitive baseline resources for Low Resource Urdu Summarization (both in terms of processing and NLP resources). The research objectives envisaged in this paper are:-
\begin{itemize}
\item To create a dataset in a low-resource language Urdu viable for the training of a summarization model not requiring specialized and expensive data collection efforts.
\item Development of a model for summarization in a low-resource language (Urdu) able to capture contextual information in a low-resource setup.
\end{itemize}

\subsection{Research Highlights}
Efforts have been made to explore pre-trained language models based on self-attentive transformer architecture for a downstream task \textit{(Automatic Summarization)} in a low-resource language. It involved exploring available datasets and models for summarization \& in the absence of a requisite dataset, creating a dataset from publicly available sources focusing on our objective to produce a competitive baseline resource for Low Resource Automatic Summarization. In summary, the main contributions of our work are as follows:

\begin{itemize}
    \item A novel methodology (mT5 $\rightarrow$ urT5), adapting pre-trained LMs based on self-attentive transformer architecture (mT5, mBERT) for low resource summarization.
    \item Creation of first summarization dataset (76.5k article, summary pairs) in a low-resource language (Urdu) from publicly available sources reproducible for other languages.
    \item Experimental results show a competitive evaluation score of 46.35 ROUGE-1 at par with state-of-the-art models despite the reduction of 44.78\% in the size of our model, with 40k monolingual vocabulary.
\end{itemize}
The rest of the paper is organized as follows; Section \ref{sec:litreview} presents literature review. Section \ref{sec:dataset} provides a detailed description of the Dataset that we have created during the course of this research. Section \ref{sec:summarization} presents the proposed methodology used in this work. The experimental setup is presented in Section \ref{sec:experiments}, followed by the results and analysis in Section \ref{sec:results}. Finally, the paper is concluded in Section \ref{sec:concl}.

\section{Literature Review}\label{sec:litreview}
Text Summarization was first focused in the late 1950s in which basic statistical or rule-based approaches were used for summarization \cite{Luhn1958} i.e. weighting sentences based on the frequency count of words. These basic statistical approaches were refined with additional features (position, cuewords, headlines, etc) and more statistical-based models \cite{Jones1972}\cite{Edmundson1969}.

Early deep learning models like RankNet \cite{Burges2005} algorithm ranked sentences as simple probabilistic cost functions using artificial neural networks. However, these models were mostly word-based models using a bag-of-words approach without catering to contextual information and dependencies in the sentence(s). Phrase clustering \cite{Lin2009} was used for NER (Named Entity Recognition) to introduce the context of words. The word "Bank" in "Bank of River" and "Bank of Punjab" have different contexts which can be differentiated using phrases instead of just words (in a bag-of-word approach). Using phrases is however an indirect method for capturing contextual information.

\textit{Rush et al.} proposed a local attention-based model \cite{Rush2015} which generates summary (headlines) by learning soft alignment between input and summary based on context. This model was inspired by \textit{Bahdanau et al.} \cite{Bahdanau2014} in which an encoder was used to encode the source sentence into a fixed-length vector. Soft alignment is used to weigh the smoothed version of source input which was used by the decoder to generate translations. A beam-search decoder \cite{Sutskever2014} is used which is a compromise between exact and greedy decoding and efficient from phrase-based machine translations in terms of computational time. This model can easily be utilized for Summarization as well by training the decoder to generate smaller (and important content) length output.

\textit{Chopra et al.} \cite{Chopra2016} extended the work of \textit{Rush et al.} \cite{Rush2015} by replacing decoder with RNNs. \textit{Hu et al.} \cite{Hu2014} and \textit{Cheng \& Lapata} \cite{Cheng2016} also made use of attentional encoder decoder RNNs inspired by \textit{Bahdanau et al.} \cite{Bahdanau2014}. \textit{Nallapati} \cite{Nallapati2016} extended the framework with addition of feature rich encoder (i.e. various feature embeddings; POS, NER, TF, and IDF are also included in encoder input in addition to word embeddings). Pointer-Generator \cite{See2017} framework used pointers for pointing source text to copy words that are required for summarization while using generators for retaining the ability to generate novel words not included in the source text. Coverage was also used to keep track of what has been summarized to discourage repetitions (shortcomings of previous RNN-based models). Hierarchical encoder \cite{Zhou2018} based on RNN used to capture document-level dependencies/context. \textit{Reinforcement Learning} \cite{Pasunuru2018} added saliency and entailment rewards for the output summary in training process.

Sequence-to-sequence models were based normally on LSTM, RNN, GRU, or CNN with an attention mechanism for capturing dependencies between tokens (i.e. context). RNN based models calculate hidden states\textsuperscript{ht} as a function of previous hidden state\textsuperscript{ht-1} for the input position\textsuperscript{t}. This sequential nature restricts parallelization within training examples putting memory constraints for longer sequences. 

Transformer model \cite{Vaswani2017} works entirely on attention mechanisms to draw global dependencies between input and output without using recurrence or convolutional networks. Multi-head attention consisting of several attention layers running in parallel based on scaled dot-product was used in Transformer Model. Self-attention was also used to reduce the computational complexity and positional encoding to keep track of sequence order. Major architectural advances made in transformer-based architecture were Parallelization and Attention Mechanism making it a suitable model for language generation with low processing resources (i.e. appropriate for Abstract Summary Generation)

Word2vec \cite{Mikolov2013}, GloVe \cite{Pennington2014} and FastText \cite{Bojanowski2017} captured latent syntactic and semantic similarities by representing words and character n-grams into vector space. ELMo (Embeddings for Language Models) \cite{Peters2018} proposed the use of Bidirectional Language Models (BiLM) for learning contextual representation for words over an input sequence utilizing LSTM based model. \textit{Zhang \& Bowman} \cite{KZhang2019} demonstrated that language modeling (LM) based pre-training objective performs better than other task-specific pre-training and also for transfer learning. For a generalized LM objective ULFiT (Universal Language Model Fine-tuning) \cite{Ruder2018} pre-trained a language model on Wikipedia articles and fine-tunes it on the downstream tasks (e.g. summarization) using novel techniques.

\textit{Transfer Learning} was enabled through learned representations of Languages by using Language Modelling (LM) and their re-use for various downstream tasks (Summarization, Q/A, Inference, etc) as depicted in Fig. \ref{fig:plm}. Transfer learning has become ubiquitous in NLP gaining popularity after the inception of transformers architecture based on attention mechanism. Previously learned word embeddings \cite{Pennington2014} \cite{Peters2018} were also used as LMs however lacking to capture long-term dependencies and contextual information. Generally LMs are pre-trained on large unlabelled data \textit{(self-supervised learning)} and then fine-tuned to target downstream tasks using labeled data.
Training of large Pre-trained Language Models was enabled with major architectural advancements and the availability of requisite processing power. These models were trained on large datasets comprising generic text with variations in pre-training strategies and size of models. The major goal of these models remained same i.e. Language Understanding and its Generation (NLU / NLG). Popular models include Bidirectional Encoder Representations from Transformers (BERT) \cite{Devlin2019}, trained on BookCorpus and English Wikipedia using masking and corruption of tokens as a pre-training strategy with encoder only transformer architecture; Text-to-Text Transfer Transformer (T5) \cite{Raffel2020} trained on Colossal Clean Crawled Corpus (C4) after analyzing various models with respect to pre-training, architectures, transfer approaches, datasets and other miscellaneous aspects which effects NLP tasks; Generative Pre-Training (GPT) \cite{Radford2018} a transformer based decoder only (unidirectional) model trained on BookCorpus with objective of predicting next token in a sentence. Initial GPT model was succeeded by a series of GPT-n Models from 2018 to 2023 with significant improvements incorporating diverse training data with GPT-2 only remaining a open-source model. The latest GPT-4 \cite{openai2023gpt4}, is a multimodal model evaluated on various real life competitive exams (e.g. Bar Exam, Chemistry, etc). Alternative open source models also became popular (BLOOM \cite{scao2022bloom}, GPT-NEO \cite{black2022gpt} with competitive evaluation scores in multiple tasks. However, challenges for Low Resource Languages still persists with issues of under-representation, biasness, non availability of requisite datasets for downstream tasks \cite{Pires2019} \cite{wang2022survey} and evaluation of such models for a low resource language etc.

\begin{figure*}[t!]
\centering
\begin{subfigure}{7.5cm}
  \centering
  \includegraphics[width=7cm]{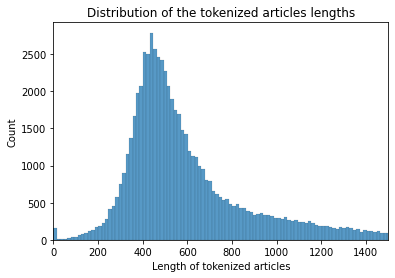}
  \caption{Articles before truncation}
  \label{fig:DWinitial}
\end{subfigure}%
\begin{subfigure}{7.5cm}
  \centering
  \includegraphics[width=7cm]{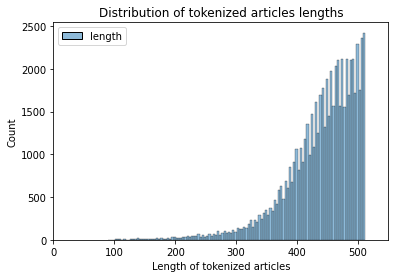}
  \caption{Articles after truncation}
  \label{fig:DWtrunc}
\end{subfigure}

\begin{subfigure}{7.5cm}
  \centering
  \includegraphics[width=7cm]{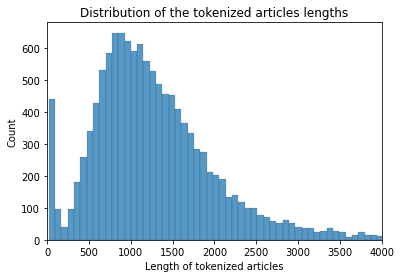}
  \caption{Articles before truncation}
  \label{fig:BBCinitial}
\end{subfigure}%
\begin{subfigure}{7.5cm}
  \centering
  \includegraphics[width=7cm]{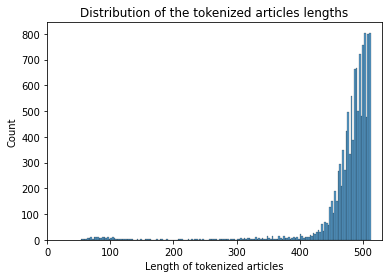}
  \caption{Articles after truncation}
  \label{fig:BBCtrunc}
\end{subfigure}

\begin{subfigure}{7.5cm}
  \centering
  \includegraphics[width=6.8cm]{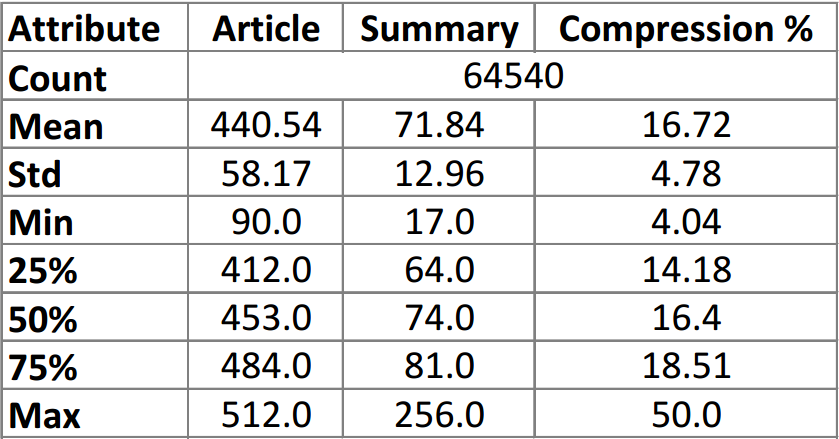}
  \caption{Tokens stats after preprocessing}
  \label{fig:DWtruncmt5}
\end{subfigure}
\begin{subfigure}{7.5cm}
  \centering
  \includegraphics[width=6.8cm]{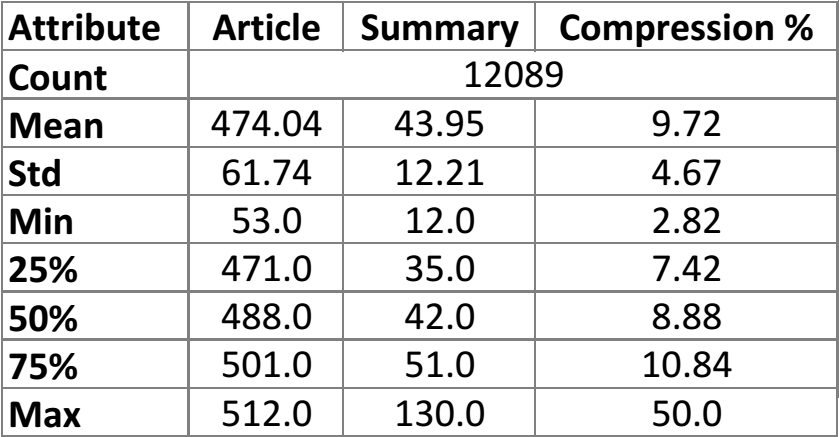}
  \caption{Tokens stats after preprocessing}
  \label{fig:BBCtruncmt5}
\end{subfigure}
\caption{Distribution of word count depicting article lengths; DW Urdu in (a) \& (b), BBC Urdu in (C) \& (d). Words statistics (based on spaCy tokenizer) of articles \& summary and compression ratio in \%; DW Urdu and BBC Urdu in (e) \& (f) respectively.}
\label{fig:dataset}
\end{figure*}

\section{Summarization Dataset}\label{sec:dataset}
Urdu is the 10th most spoken language\cite{Ethnologue}, (230 million people) in the world. It is a member of the Indo-Aryan group in the Indo-European family of languages. It is widely spoken in South Asian regions, Pakistan, India, Afghanistan, Bangladesh, Nepal and Bahrain. Its vocabulary is derived from Persian, Arabic, and Turkish. Urdu also shares its origins with the Hindi language however Hindi is written in Devanagari, the same script as Sanskrit, and its vocabulary has more of a Sanskrit influence than Persian and Arabic Influence. If spoken colloquial contexts are broadly considered, (Hindi-Urdu) is the 3rd most spoken language in the world.

Urdu is a relatively complex and morphologically rich language. Urdu script is written in Nastaliq style in which most of the characters acquire different shapes depending on the position of the character in the ligature. It is written from right to left and blank spaces don't necessarily mean the segregation of words as in the English language, hence word and sentence boundary detection are difficult in Urdu. Moreover, there is no concept of word capitalization in Urdu making tasks like NER, and sentence segmentation by detecting boundaries through capitalization becomes more difficult. Despite being a popular language with a lot of content over the internet in multiple forms (books, articles, news, microblog posts, forums, etc) there is little to no research related to Urdu language available in the field of Automatic Summarization and NLP.

\subsection{Dataset Creation \& Preprocessing}\label{ss-datasetpp}
To the best of our knowledge, only one summarization dataset is available in Urdu language \cite{Humayoun2016} having only 50 records, hence making it infeasible for training machine learning algorithm for summarization. Recent work in Urdu Language includes emotion detection with annotated dataset \cite{bashir2022context} and Extractive Summarization comparing various word/term based methods \cite{nawaz2020extractive} using the same dataset \cite{Humayoun2016} comprising 50 records. The creation of a requisite large human-written document/summary dataset is an expensive and time taking task. Recently datasets have been created by utilizing the existing resources from publicly available data \textit{(i.e. reviews and their summaries, news websites containing article / summary pairs, etc)} \cite{Narayan2018}. The news domain is a suitable choice for the creation of summarization dataset owing to:-
\begin{enumerate}
    \item Publicly available.
    \item Easily collectible.
    \item Available in multiple/local languages.
    \item Originally written by multiple human authors contrary to synthetic dataset approaches.
\end{enumerate}
Availability of popular news platforms (in addition to local news resources) in multiple languages also makes it the best possible choice for the acquisition of summarization dataset.

Two news websites (BBC Urdu\footnote{BBC Urdu - \url{https://www.bbc.com/urdu}} and DW Urdu\footnote{DW Urdu - \url{https://www.dw.com/ur}}) were selected which have 2-3 lines of short summary written by multiple writers in addition to the news articles. These two websites were scrapped for article/summary pairs. A dataset of 76.5k data points having Article/Summary pairs were scrapped \textit{(12k from BBC Urdu and 64.5k from DW Urdu)}. The dataset was tokenized using spaCy tokenizer\footnote{spaCy - \url{https://spacy.io/}} (word-based tokenizer), mBERT and mT5 (upto sub-word tokenizer; i.e. WordPiece) for exploration of dataset and length analysis. Detailed statistics of tokenized lengths and compression ratio (before / after preprocessing) are shown in Fig. \ref{fig:dataset}.

Preprocessing steps involved are:-
\begin{itemize}
\item \textit{Multimedia} - Only text-based articles were selected to be included in the dataset excluding Multimedia Based Articles.
\item \textit{Links / URLs} - All types of links, URLs were removed. For instance, links of associated articles were removed from articles.
\item \textit{Picture Captions} - These news websites also had pictures, screenshots of tweets, etc. inside article which also had captions, picture captions were removed while scrapping.
\item \textit{Compression Ratio} - Compression Ratio was calculated for each record using tokenized length. Records having compression ratio more than 50\% were removed \textit{(i.e. 830 records)}. 
\end{itemize}

\begin{figure}[h]
    \centering
    \includegraphics[width=8.5cm]{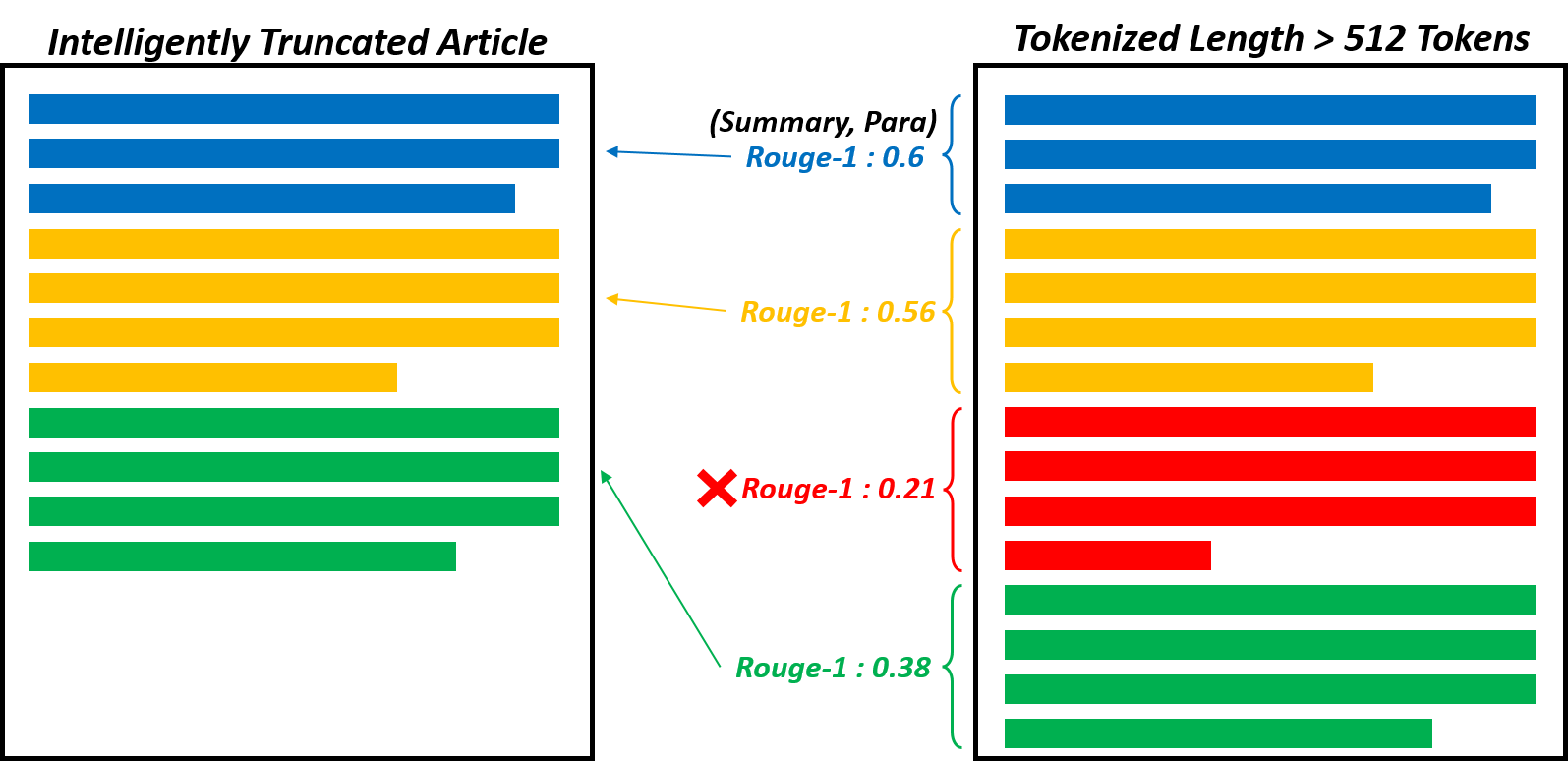}
    \caption{Intelligent Truncation using Recall between article \& summary paras.}
    \label{fig:trunc}
\end{figure}

\renewcommand{\algorithmicrequire}{\textbf{Input:}}
\renewcommand{\algorithmicensure}{\textbf{Output:}}
\renewcommand{\algorithmicforall}{\textbf{for each}}
\floatname{algorithm}{Procedure}
\begin{algorithm}[h]
\caption{Truncation of Dataset using Recall Measure}\label{algo}
\begin{algorithmic}
\Require $article, summary$
\Procedure{Tokenized\_Length}{$article$}
	\State $Encode\_Articles$ \Comment{e.g.using BERT Tokenizer}
	\ForAll{$para \in article$}
		\State $length_{para} = len(para \in tokenized\_article)$
	\EndFor
	
	\Return $length$
\EndProcedure
\If{$length_{article} > 512$}
\Procedure{Score\_Paragraphs}{$article, summary$}
	\ForAll{$para \in article$}
		\State $para_{index} = i+1$
		\State $para_{text} = para$
		\State $Rouge\_Score(article, summary)$
		\State $para_{score} = Rouge-1_{Recall}$
	\EndFor
\EndProcedure
\EndIf
	\While{$length_{article} > 512$}
		\State $sorted\_paras = SortAsc(para_{score})$
		\State $i = 0$
		\State del $sorted\_paras_i$
		\State $length_{article}  -= length_{{para}_i}$
		\State $i += 1$
	\EndWhile
\State $trunc\_article = SortAsc(para_{index})$
\Ensure $trunc\_article$
\end{algorithmic}
\end{algorithm}

\subsection{Dataset Truncation}\label{ss-trunc}
Pre-trained language models based on BERT used in extractive summarization in section \ref{ss-extsum} limits processing of input documents up to 512 tokens. These model automatically discards the remaining text of the input document resulting into the loss of important information. Articles in both datasets had more then 512 tokens which will result in loss of information useful for summarization. Therefore, Intelligent Truncation has been carried out using Recall measure between article text paragraphs and summary text to cater for the limitations of BERT-based models. Model mT5 used for Abstractive Summarization in section \ref{ss-abssum} theoretically doesn't have an input processing limit however its memory utilization exponentially increase with large input text therefore only truncated dataset up to 512 tokens was used. Truncation is explained in Procedure \ref{algo} \& Fig. \ref{fig:trunc}.

\begin{figure*}[h]
    \centering
    \includegraphics[width=\textwidth]{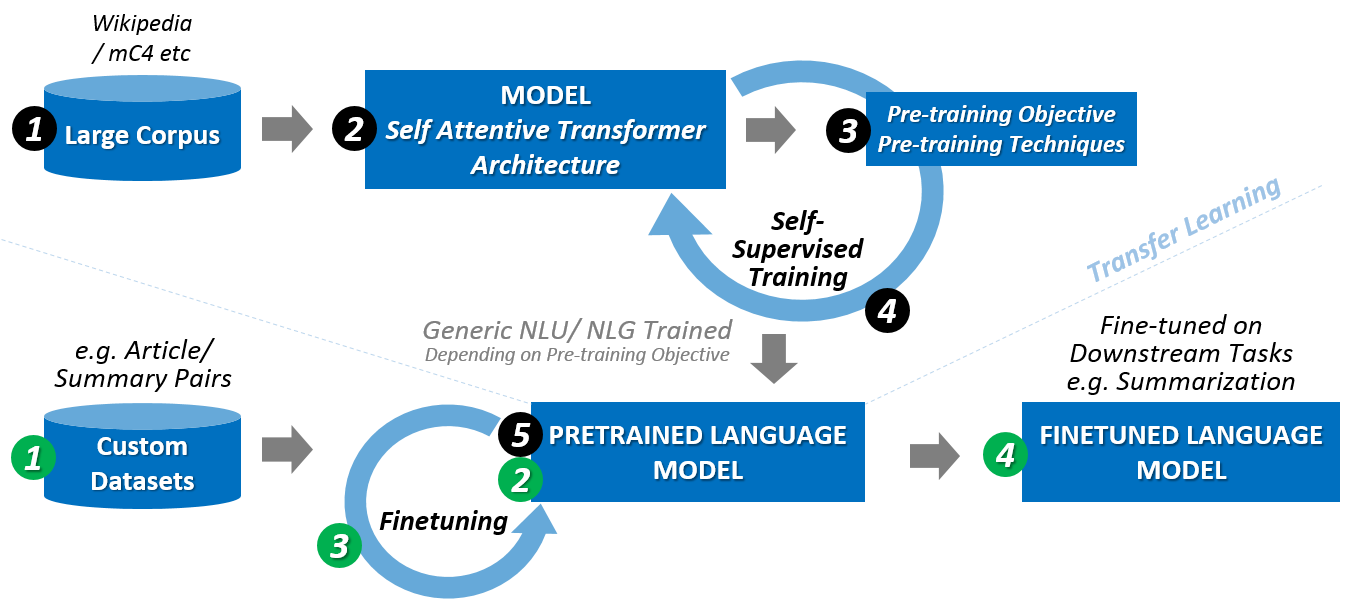}
    \caption[Transfer Learning; Language Models]{Pre-Trained Language Models \textit{(Transfer Learning; re-useable Language Models)} \& Fine-Tuning for downstream tasks}
    \label{fig:plm}
\end{figure*}
\section{The Proposed Framework: Summarization using Pre-trained Language Models}\label{sec:summarization}
Pre-trained Language Models \textit{(depicted in Fig. \ref{fig:plm})} involves training of a model over a large corpus to learn the ability to understand and generate language representations for downstream tasks (summarization etc). Low resource languages (Urdu; selected in this research) suffers from lack of available pre-trained language models. Pre-training objective previously confined to monolingual was extended to multilingual training \cite{Devlin2019}, \cite{Artetxe2018}\cite{Lample2019}. Evaluation of these multilingual models on various tasks showed improved results. Various popular models e.g. BERT \cite{Devlin2019}, T5 \cite{Raffel2020} \cite{Xue2020} etc have been released with a scaled objective of multilingual pre-training. Multilingual models are trained in parallel over large corpus of multiple languages (BERT \& T5 used in this research trained over more than 100 languages) with shared vocabulary. This combined training though suffers from under-representations of language (low resource) having comparatively less training data \cite{Pires2019} but still provides workable language model which can be used for various downstream tasks efficiently. Various models also explored cross-lingual training with parallel training data with high resource languages through various automatic techniques (e.g. machine translation of training data in English language to a low resource language) creating synthetic datasets to cater for problem of low resource languages however restricted to only few languages.

\begin{figure*}[h]
    \centering
    \includegraphics[width=\textwidth]{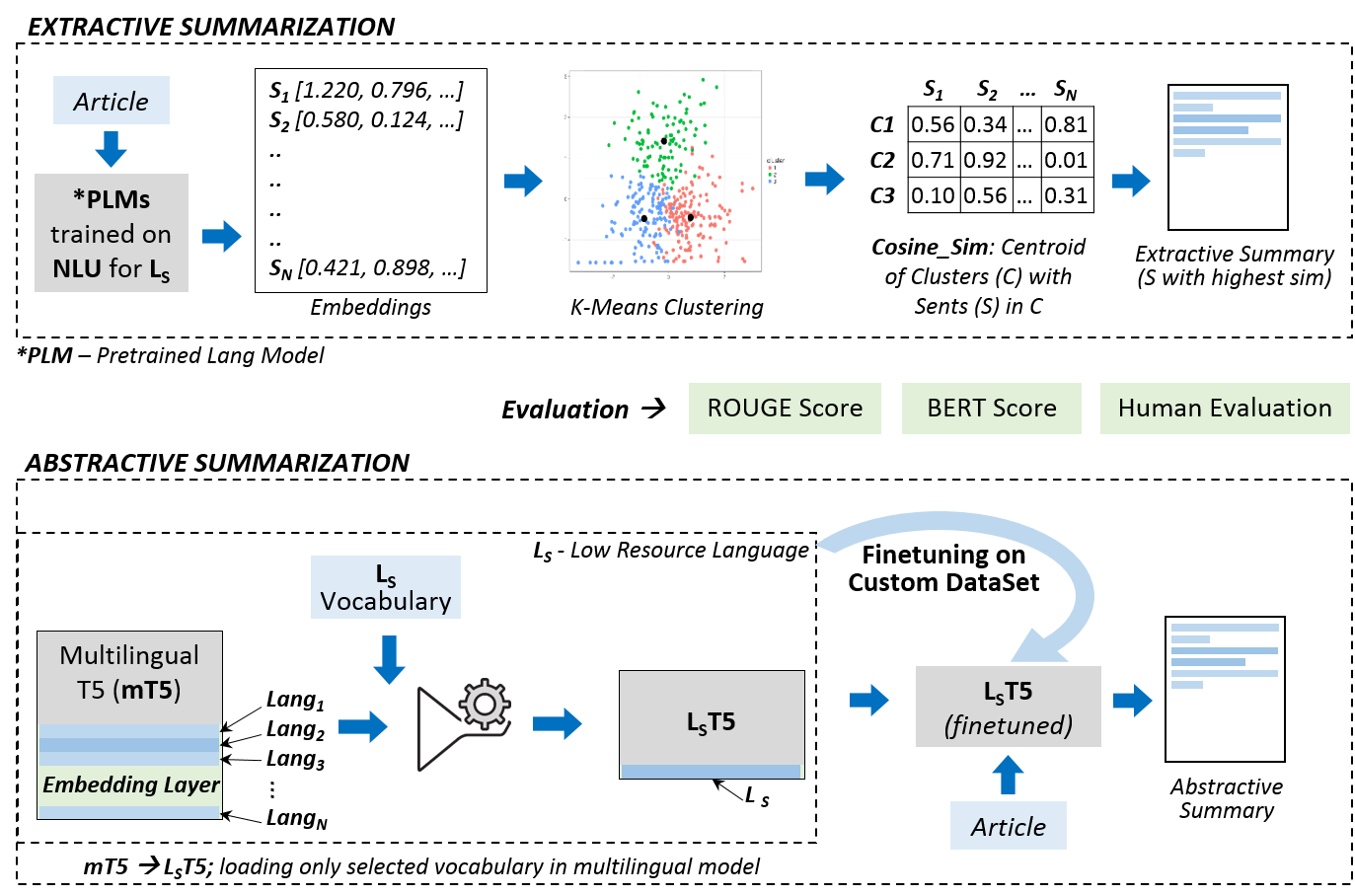}
    \caption[Summarization Framework]{Adopted Summarization Framework}
    \label{fig:framework}
\end{figure*}

Summarization has been carried out in both extractive and abstractive category by first creating own dataset as described in section \ref{sec:dataset}, later by choosing a viable multilingual pre-trained language model which is also trained in our selected low resource language (i.e. Urdu) and its adaptation by reducing its size to fit in low resource settings. Methodology adopted to carry out our research is illustrated in Fig. \ref{fig:framework} and explained in section \ref{ss-extsum} and \ref{ss-abssum}. Publicly available models selected for the research are Multilingual BERT \textit{(mBERT)}, Multilingual T5 \textit{(mT5)}, Multilingual Representations for Indian Languages; a BERT based model \textit{(MuRIL)}. These models are being used in following portion of paper for summarization using own created dataset. \textit{mBERT based models are used for extractive summarization and mT5 with generative capability for abstractive summarization}.

\subsection{Extractive Summarization using BERT based models}\label{ss-extsum}
BERT (Bidirectional Encoder Representations for Transformers) \cite{Devlin2019} used masked language as a pre-training objective using unlabelled data \textit{(i.e. self-supervised training)} for NLU. It introduced the concept of bidirectional pre-training unlike previous work in which either unidirectional language models pre-training was used i.e. GPT (Generative Pre-Training) \cite{Radford2018} or concatenation of independently trained left-to-right and right-to-left LMs. BERT architecture was based on Self-Attentive Transformers \cite{Vaswani2017} based encoder only model.

Extractive Summarization has been carried out on both Truncated (as explained in \ref{ss-trunc}) and Non-truncated datasets. Extractive Summarization has been carried out using mBERT-based models which are trained for NLU. These mBERT-based models are used as zero-shot settings without finetuning for finding the contextual embeddings, a task similar to the pre-training objective of these models i.e. NLU. Sentences \(\{S_j\}_1^N\) in a Summary are encoded by these models into Sentence Embeddings \(E_{Sj}\). Embeddings were clustered into k clusters \(\{C_i\}_1^k\) using k-means clustering. Sentence closest to the Centroid \(O_i\) was selected as a Summary Sentence \(X_i\) based on Cosine Similarity measure \(Cosine\_Sim(O_i, E_{Sj})\). These summary sentences were included in the final summary in the order in which they occurred in the article. No of clusters k is found by Number of Sentences N in a Summary \(\times\) Ratio of Tokens in Article A and Summary Y.
\begin{equation*}
    \begin{split}
    & k = Tokens(Y)/Tokens(A) \times N \\
    \forall & C_i = \{C_1, \dots, C_k\}; X_i = \{S_{C_1}, \dots, S_{C_k}\} \\
    \forall & E_{Sj} = \{E_{S1}, \dots, E_{SN}\} \in C_i; \\
       & S_{C_i} = \max(Cosine\_Sim(O_i, E_{Sj}))
    \end{split}
\end{equation*}

Details of mBERT-based models used for extractive summarization are:-
\begin{itemize}
    \item mBERT - trained over 104 languages; base - 110M parameters, $\approx681M$ size
    \item MuRIL \cite{Khanuja2021} - trained over 17 Indian languages; base - 236M parameters, $\approx909M$ size \& large $\approx1.89G$ size
    \item Geotrend/BERT \cite{Abdaoui2020} - monolingual Urdu version of mBERT with 48\% reduced size i.e. $\approx354M$ and reduced memory utilization considering low resource settings
\end{itemize}

\subsection{Abstractive Summarization using adapted mT5 $\rightarrow$ urT5}\label{ss-abssum}
T5 (Text-to-Text-Transfer-Transformer)\cite{Raffel2020} goal was not to propose a new model of innovative architecture. Instead, a deep study of all available research including pre-training, architectures, datasets, transfer approaches, and other miscellaneous aspects which affects NLP tasks was carried out and efficient approaches were selected for the creation of a text-to-text model. It comprised of a survey of existing research in the field, their comparison, limitations, and in the end utilizing the takeaways of the study and training a model which achieves state-of-the-art (SOTA) in language understanding \& generation and various downstream tasks (e.g. summarization). The baseline model was designed with an encoder and decoder. The encoder/decoder architecture was similar to BERT \cite{Devlin2019} (except that BERT was an encoder-only model). The model was trained using "masked language modeling" and denoising objective inspired by BERT. Zeroshot settings of the model may not provide acceptable results on summarization as it is trained for understanding \& generating language representations only instead of specific tasks like summarization. With little training (i.e. finetuning) models can outperform various old methods of summarization (e.g. statistical). Specifically language generation capability of the T5 model is exploited for abstractive summarization which was not possible with earlier models.

Adapted mT5 (Multilingual T5) \cite{Xue2020} was selected for Abstractive Summarization which is trained over mC4\footnote{Multilingual C4 - \url{https://www.tensorflow.org/datasets/catalog/c4}} covering 102 languages following similar recipe as T5. mT5 has a pre-training objective of NLU \& NLG but this generation capability is not particularly trained for generating summaries and hence needs finetuning. 

\subsubsection{Adaptation of Multilingual T5 to Urdu T5 (mT5 $\rightarrow$ urT5)}
mT5 has 5 checkpoints (small, base, large, XL, XXL). Due to the extensive size of large checkpoints requiring requisite memory during finetuning, mT5-small \& mT5-base were selected for experimentation. These multilingual models can be more efficiently utilized for monolingual tasks by reducing the shared vocabulary of the model to monolingual vocabulary as proposed by \textit{Abdaoui et al.} in Geotrend/BERT models \cite{Abdaoui2020}. \textit{Abdaoui et al.} proposed loading only monolingual vocabulary in a multilingual model as most of the parameters of these multilingual models are in embedding layers. By reducing the vocabulary; input/output embeddings of the model are reduced. As a result size of the model and its memory utilization is reduced retaining almost the same efficiency as of original multilingual model. Monolingual vocabulary used in our adapted model comprised of 40k tokens collected from \textit{1M Urdu News Dataset\footnote{1M Urdu News Classification Dataset - DOI: 10.17632/834vsxnb99.3 \url{https://data.mendeley.com/datasets/834vsxnb99/3}}} \& own created dataset as compared to 250k tokens of mT5-base (also explained in Fig. \ref{fig:framework}). As a result, the size of the monolingual model (we called it urT5) adapted from multilingual model mT5 was reduced to 44.78\% of its original size \textit{(mT5-base:2.17GB $\rightarrow$ urT5-base:1.04GB)}.

\section{Experiments}\label{sec:experiments}
Experimentation was carried out using Google’s Colaboratory (Colab) platform being accessible freely without a specialized environment setup (achieving the aim of low resource summarization). Google’s Colab is a free development environment based on a Jupyter Notebook environment that runs on a cloud supporting collaborative developments. It supports popular ML libraries and offers a limited amount of GPU (i.e. $\approx12GB$). Free usage has other limitations including session usage time, inactivity time, background execution limitations, etc.

\subsection{Evaluation Methodologies}\label{ss-eval}
\subsubsection{ROUGE}\label{sss-rouge}
Evaluation of summarization is normally carried out through Recall-Oriented Understudy for Gisting Evaluation (ROUGE) \cite{Lin2004}. The main idea of ROUGE is to calculate terms that overlap between the original gold summary which is normally written by humans and generated or predicted summary by the model. The basic evaluation measure in ROUGE is ROUGE-N in which N is N-gram overlap statistics including Precision, Recall, and F-Measure. ROUGE Evaluation has inherent issues not restricted to preprocessing steps involved before the evaluation phase but also to abstractive summarization. Abstractive summarization includes words/phrases which are not included in original gold/reference summaries but are generated innovatively to fit into the context of the sequence being generated. ROUGE evaluation depending upon N-gram co-occurrences/ overlap becomes contrary to the very concept of abstractive summarization. Moreover, it considers a sequence as bag-of-words that takes out contextual information and its dependencies over the complete sequence. There may be cases where a summary is evaluated as a good quality with a high evaluation score however in human evaluation it may score as inferior and vice versa.

\subsubsection{BERTScore}\label{sss-bertscore}
Most of the evaluation methods proposed earlier were based on the exact matching of N-grams like ROUGE for summarization, METEOR (Automatic Machine Translation Evaluation System), BLEU (Bilingual Evaluation Understudy) for machine translation, etc. After the release of pre-trained language models which were successfully demonstrated to capture contextual information in sequence(s). To overcome the shortcomings of exact word matching, BERTScore \cite{Zhang2019} was introduced recently which instead of exactly matching the N-grams, calculates the similarity between the contextualized token embeddings. By considering the similarity between contextual token embeddings, paraphrasing as well as dependencies between words were also catered for, which was not considered by metrics similar to ROUGE. This improvement in the contextual aspects of evaluation doesn't necessarily mean BERTScore will correctly identify the high-quality summaries due to its inherent dependency on the BERT model and its learning of language representation along with its inherent shortcomings.

\subsubsection{Human Evaluation} \label{sss-humEval}
To overcome the impediments of automated evaluation (as discussed earlier in section \ref{sss-rouge} \& \ref{sss-bertscore}) and the absence of a unanimous standard. Few generated summaries were evaluated by humans which had their primary language as Urdu. Human evaluation has been carried out on 20 x summaries generated by each model/dataset (Section \ref{sec:results}) and its comparison with automated evaluation metrics in Fig. \ref{fig:EvalComparison}. To correctly validate the evaluation results of already used metrics and to verify the quality of summaries, various summaries have been selected; ranked higher, lower, and in mid-range.

To create the evaluation process easy only gold reference summaries and generated summaries were presented to the evaluators to avoid readers' biases and also prevent their disinterest in reading long articles. Ranking of the summaries has been carried out considering two factors on the scale of 0 \textit{(considered as Lowest)} to 5 \textit{(considered as Highest)} assuming the reference summary as the gold standard and true in all aspects:-
\begin{itemize}
\item Accuracy / Relevance - Information conveyed in the summary predicted by the model is accurate, consistent, and relevant as conveyed in the original reference summary written by the human author.
\item Coherence - Ability to convey information with continuity and linked ideas and language together to form coherent, well-formulated, and connected sentences; as conveyed in original reference summary written by the human author.
\end{itemize}

\begin{table}[ht]
\centering
\renewcommand{\arraystretch}{1.2}
\captionsetup{font=small}
\caption{Extractive Summarization: BBC Urdu Dataset (R-1; ROUGE-1 F Score and B-S; BERT Score)}
\begin{tabular}{m{3.1cm} | m{0.8cm} m{0.8cm} m{0.8cm} m{0.8cm} } \hline 
Model & R-1 & R-2 & R-L & B-S \\ \hline 
mBERT-base & 39.59 & 23.50 & 33.30 & 74.59 \\ \hline 
mBERT-base (Trunc) & 47.98 & 31.59 & 41.91 & 77.61 \\ \hline 
MuRIL-base & 39.30 & 23.42 & 33.21 & 74.33 \\ \hline 
MuRIL-base (Trunc) & 47.03 & 30.77 & 40.98 & 77.16 \\ \hline 
MuRIL-large & 40.73 & 24.35 & 34.37 & 74.95 \\ \hline 
MuRIL-large (Trunc) & 48.75 & 32.35 & 42.63 & 77.83 \\ \hline 
Geotrend-BERT-base & 39.58 & 23.46 & 33.28 & 74.57 \\ \hline 
Geotrend-BERT-base (Trunc) & 47.99 & 31.59 & 41.90 & 77.60 \\  \hline
\end{tabular}
\label{table:r-ExtBBC}
\end{table}

\begin{table}[ht]
\centering
\renewcommand{\arraystretch}{1.2}
\captionsetup{font=small}
\caption{Extractive Summarization: DW Urdu Dataset (R-1; ROUGE-1 F Score and B-S; BERT Score)}
\begin{tabular}{ m{3.1cm} | m{0.8cm} m{0.8cm} m{0.8cm} m{0.8cm} } \hline 
Model & R-1 & R-2 & R-L & B-S \\ \hline 
mBERT-base & 30.62 & 9.82 & 21.13 & 71.52 \\ \hline 
mBERT-base (Trunc) & 34.19 & 12.18 & 23.82 & 72.63 \\ \hline 
MuRIL-base & 30.19 & 9.64 & 21.15 & 71.50 \\ \hline 
MuRIL-base (Trunc) & 33.29 & 11.66 & 23.40 & 72.23 \\ \hline 
MuRIL-large & 30.95 & 9.98 & 21.46 & 71.6 \\ \hline 
MuRIL-large (Trunc) & 34.09 & 12.24 & 23.94 & 72.60 \\ \hline 
Geotrend-BERT-base & 30.59 & 9.80 & 21.01 & 71.50 \\ \hline 
Geotrend-BERT-base (Trunc) & 34.23 & 12.22 & 23.85 & 72.64 \\ \hline
\end{tabular}
\label{table:r-ExtDW}
\end{table}

\begin{table}[h!]
\centering
\renewcommand{\arraystretch}{1.2}
\captionsetup{font=small}
\caption{Abstractive Summarization (F; F Score, P; Precision ROUGE-1 \& B-S; BERTScore)}
\begin{tabular}{ m{3.6cm} | m{1cm} m{1cm} m{1cm} } \hline 
Model & F & P & B-S \\ \hline 
urT5-base (w/o fine-tune) & 19.54 & 21.77 & 58.42 \\ \hline
mT5-small & 36.43 & 37.37 & 73.36 \\ \hline 
urT5-small & 36.39 & 37.41 &  73.43 \\ \hline 
urT5-base & 39.92 & 44.14 &  75.07 \\ \hline
urT5-base (50\% epochs) & 40.03 & 44.32 &  75.1 \\ \hline 
urT5-base (50\% dataset) & 39.13 & 43.47 &  74.77 \\ \hline 
urT5-base (50\% dataset + 50\% epochs) & 38.03 & 42.64 & 74.27 \\ \hline
urT5-base BBC Urdu & 46.35 & 52.12 & 77.0 \\ \hline
urT5-base DW Urdu & 36.91 & 40.4 & 74.17 \\ \hline
\end{tabular}
\label{table:r-Abs}
\end{table}

\section{Results and Analysis}\label{sec:results}
\textit{Extractive Summarization} was performed on BBC Urdu and DW Urdu datasets separately, results of which are shown in Table \ref{table:r-ExtBBC} and \ref{table:r-ExtDW}. Extractive Summarization achieved up to 48.75 ROUGE-1 (largest model; MuRIL-large $\approx1.89GB$) and 47.99 ROUGE-1 (smallest model; Geotrend-BERT-base $\approx354MB$) on BBCUrdu dataset however comparatively lower scores (34.23 ROUGE-1 of Geotrend-BERT-base) in DWUrdu dataset due to comparative complexity and abstractive features of DWUrdu dataset. Equivalent scores of smaller sized model with lower resource consumption verify our research objectives.

For \textit{Abstractive Summarization}, variations of our adapted models based on mT5 were trained on a dataset of 72k comprising of combined BBC Urdu and DW Urdu datasets. The training was carried out up to 5 epochs with a batch size of 4 \& gradient accumulation of 8. Testing was carried out on joint BBC, and DW Urdu datasets as well as on separate subsets. Results are shown in Table \ref{table:r-Abs}; urT5-base $\approx1.02GB$ has +3.6 ROUGE-1 improvement as compared to its comparative sized model mT5-small $\approx1.2GB$ whose training/ finetuning could fit in Google's Colab \textit{(publicly available free version)}.

\subsection{Effects of Intelligent Truncation}
Evidently, 512 truncated versions of both datasets have high evaluation scores in \textit{extractive summarization} Table \ref{table:r-ExtBBC} \& \ref{table:r-ExtDW}. In \textit{abstractive summarization} 512 truncated version was only used both for training and testing to cater for the memory utilization of mT5-based models during training in our selected low resource setup.

\subsection{Low Resource Models; Geotrend/BERT \& mT5 $\rightarrow$ urT5}
Due to the absence of monolingual models for low-resource languages, multilingual models are a suitable alternative to be used for monolingual purposes. This task was achieved in an efficient manner by reducing the size \& memory requirements of models. In both \textit{extractive and abstractive summarization} no degradation in the evaluation of adapted models is observed as compared to their actual multilingual models despite a significant reduction in size. If compared with similar-sized models, adapted models showed up to +3.6 ROUGE-1 improvement in abstractive summarization (Table \ref{table:r-Abs}).



\begin{figure*}[ht]
    \centering
    \includegraphics[width=\textwidth]{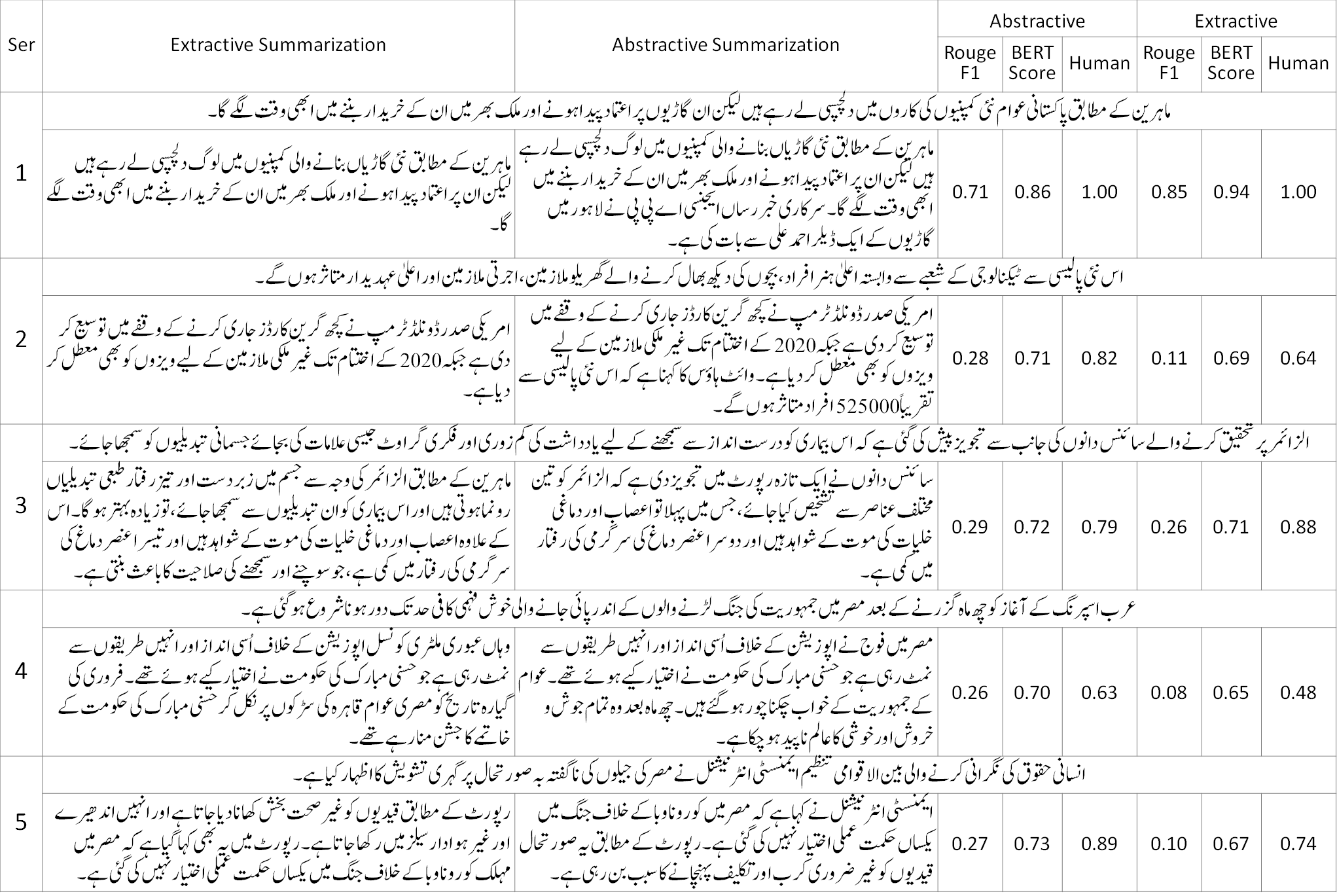}
    \caption{Selected samples from summaries undergone human evaluation; Actual Summary in top merged row, Extractive \& Abstractive Summarization and Evaluation Scores in bottom row}
    \label{fig:selEval}
\end{figure*}

\subsection{Extractive vs Abstractive Summaries}
In \textit{extractive summarization} evaluation score of the BBC Urdu dataset is comparatively high as compared to that of the DW Urdu dataset. Summaries with top evaluation scores were analyzed and found that BBC Urdu has a number of summaries having maximum terms extracted from articles hence increasing term overlap resulting in highly automated evaluation scores. In \textit{abstractive summarization} training was carried out by a joint dataset however the same effect was observed while evaluating BBC Urdu and DW Urdu separately. BBC Urdu evaluation score was comparatively high however effect was reduced as compared to extractive summarization results because of the capability of mT5-based models to generate abstract summaries instead of selecting sentences from input text.

\subsection{MuRIL used for Extractive Summarization}
mBERT is trained over 104 languages including language under research i.e. Urdu with a large corpus of Wikipedia. MuRIL \cite{Khanuja2021} is a BERT-based model pre-trained on 17 Indian languages which contained translated and transliterated documents as well for cross-lingual training from Wikipedia, Common Crawl, PMINDIA, and Dakshina. MuRIL was used for \textit{extractive summarization only} whose evaluation score shows no improvement (base versions) as compared to mBERT which is only 75\% of its size (Table \ref{table:r-ExtBBC} \& \ref{table:r-ExtDW}) mainly due to quality aspects (translated and transliterated text in training dataset etc).

\subsection{Training Aspects in Abstractive Summarization (Table \ref{table:r-Abs})}
\begin{itemize}
\item \textit{Dataset} - A larger dataset with more training examples improves the model's ability to summarize which is already well known. However, fewer training examples up to a threshold should be sufficient for satisfactory results for low resource summarization as a minor difference in evaluation is observed as compared to the reduction in training data (i.e. 0.89 for reducing 30k training data from 72k). Aspects including dynamic and quality training data samples need exploration in addition to a number of training data points.
\item \textit{Training Epochs} - Fine-tuning has been carried out for 5 epochs however evaluation has also been carried out for 2.5 epochs (50\%). Though there is a minor difference in evaluation scores it was observed that more training doesn't necessarily mean high evaluation or efficiency of the trained model (overfitting etc).
\item \textit{Abstractive vs Extractive} - Automated evaluations are usually comprised of term overlap or contextual similarity of terms overlap. In relatively simple summaries, the extractive nature is favorable for high scores in currently adopted automated evaluation metrics. However in reality datasets are more complex in nature which favours for abstractive summarization to convey information presented in a complex input text. Inherent limitations of automated evaluation metrics and their direct effect on loss function in training become a hindrance in developing models which can understand complex text (e.g. sarcasm, idioms, etc).
\item \textit{Zeroshot Evaluation} - mT5-based models are trained for NLG but not particularly for summary generation tasks therefore as expected results of usage of these models without finetuning are quite low.
\end{itemize}

\begin{figure*}[h]
    \centering
    \includegraphics[width=\textwidth]{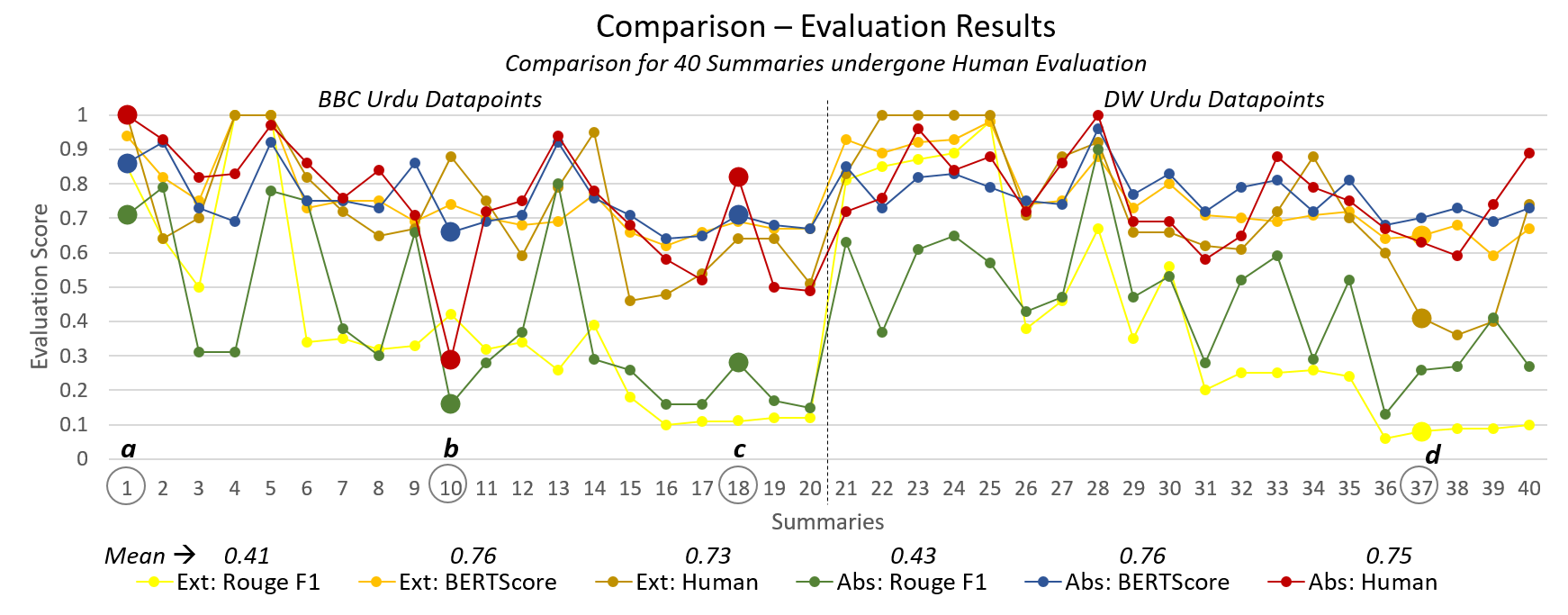}
    \caption{Comparison of Human Evaluation Scores of 40 selected summaries; Rouge-1 F Score, BERTScore \& Human Evaluation for Extractive \& Abstractive Summarization. a. Summaries ranked 100\% by human evaluators ranked comparatively lower by automated evaluation metrics. b. High BERTscore only due to high semantic similarity of words although summary failed to convey desired information; ranked lower by human \& Rouge. c. Summaries with different words but the same meanings (abstractive) ranked with high BERTScore \& human evaluation but ranked lower by Rouge. d. Variation in evaluations due to complex concepts like cohesiveness, quality, usefulness, etc.}
    \label{fig:EvalComparison}
\end{figure*}

\subsection{Comparison with Human Evaluation}
Human Evaluation has been carried out to cater to the deficiencies of automated evaluations and to verify the results and findings of the experimentation. Evaluation has been performed by 10 Human evaluators whose primary language is Urdu and who are qualified to understand the criteria set-up for the evaluation of summaries as described in section \ref{sss-humEval}. Results of the evaluation score of 20 summaries from each dataset are shown in Fig. \ref{fig:EvalComparison}. Evaluation has been carried out for the same summaries for extractive as well as abstractive summarization to draw a fair comparison. Major findings after carrying out human evaluation on a few summaries are \textit{(Fig. \ref{fig:EvalComparison} \& Fig. \ref{fig:selEval})}:-
\begin{itemize}
\item \textit{\textbf{Best Summaries Unanimously}} - Simpler text and their summaries were mostly graded with high scores (no distinguishable differences) by automated evaluation metrics (ROUGE \& BERTScore) as well as human evaluation.

\item \textit{\textbf{Automated Evaluations; Word Count \& Term Matching / Semantic Similarity}} - Best Summaries as per human evaluators were also ranked comparatively lower by BERTScore and ROUGE due to the statistical approach of total word count and Term Matching / Semantic Similarity of Words. \textit{(Summary 1 in Fig. \ref{fig:selEval}, visual interpretation in Fig. \ref{fig:EvalComparison}; BBC Urdu Results - Summary 1).}

\item \textit{\textbf{BERTScore; Term based Semantic Similarity}} - Summaries with lack of cohesiveness and failing to convey the same information as actual summary were ranked lower by human evaluators and ROUGE (lesser word overlap) but BERTScore ranked these summaries higher only due to high semantic similarity of words. \textit{(Summary 3 in Fig. \ref{fig:selEval}, visual interpretation in Fig. \ref{fig:EvalComparison}; Abs Score of Summary 10 in BBC Urdu Results).}

\item \textit{\textbf{ROUGE \& Generative Models}} - Abstractive Summaries by generative models have new words having similar meanings which are not present in the actual summary. Such summaries tend to be ranked lower by ROUGE but higher by both BERTScore and human evaluators. \textit{(Summary 2 in Fig. \ref{fig:selEval}, visual interpretation in Fig. \ref{fig:EvalComparison}; Abs Score of Summary 18 in BBC Urdu Results).}
 
 \item \textit{\textbf{Variance in Evaluations}} - Few concepts like cohesiveness, quality, and usefulness of Summaries cannot be captured by current evaluation metrics. Considering these complex attributes, human evaluators tend to rank summaries differently from automated evaluations. Variations also exist between ROUGE and BERTScore. \textit{(Summary 4 in Fig. \ref{fig:selEval}, visual interpretation in Fig. \ref{fig:EvalComparison}; DW Urdu Results - Summary 17, Ext Score).}

\end{itemize}

Human Evaluation also has biases/differences of opinion for the standard of summary. However neglecting the issue of personal preferences, considering the current models and evaluation methods one can have a fair idea that certain types of complex text may not have accurate automatic evaluation scores requiring additional evaluations.

\begin{figure*}[h]
    \centering
    \includegraphics[width=\textwidth]{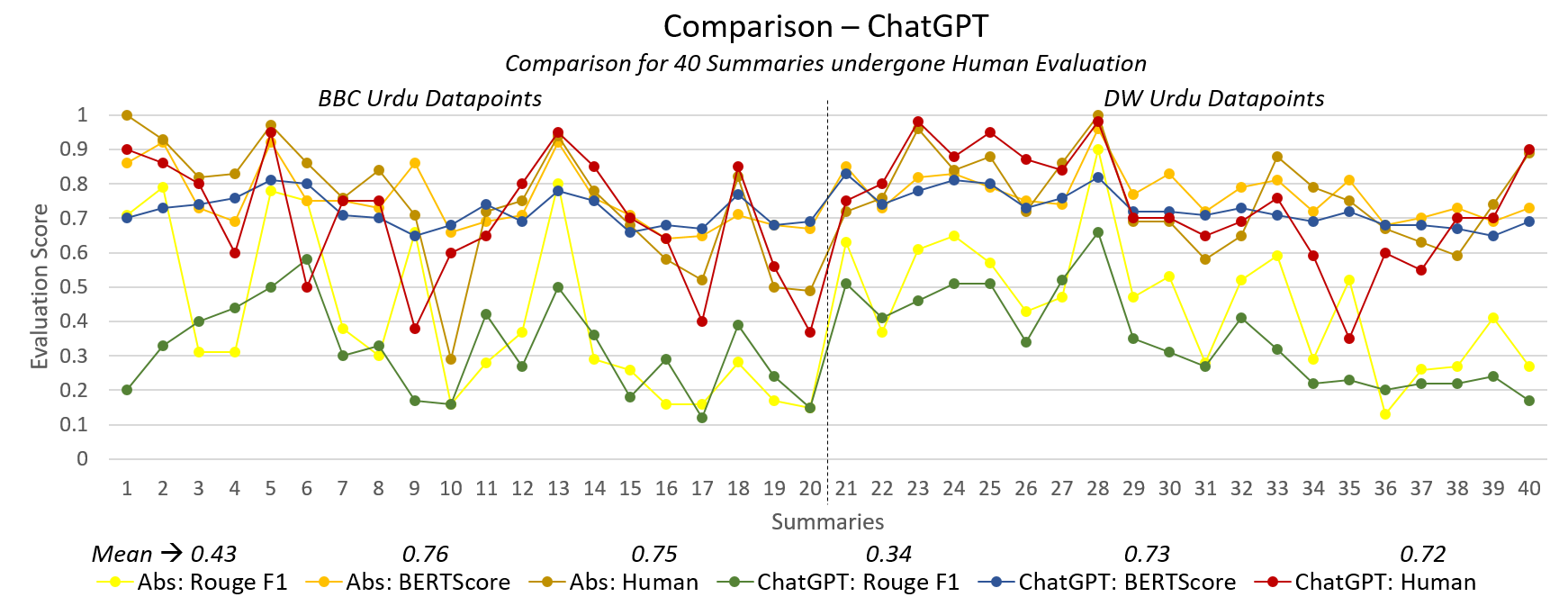}
    \caption{Comparison with popular ChatGPT model (without fine-tuning in zero-shot settings; as its not open source nor fulfills our objective of being low resource model)}
    \label{fig:ChatGPTComparison}
\end{figure*}

\subsection{Comparison with ChatGPT}
ChatGPT model is made available online through OpenAI API on 30 Nov 2022. ChatGPT has become well known for its human-like (dialogue/chat) based language generation ability in a very short period of time in a vast domain from financial analysis \& code generation to poems \& book writing etc. 
Despite its popularity and intelligent generative ability, it suffers from hallucinations \cite{ji2023survey}, factual inaccuracy \cite{bang2023multitask}, sensitive to input phrases \footnote{\url{https://openai.com/blog/chatgpt}}, fails on basic maths tasks \cite{bang2023multitask} etc yet perform complex code analyses. Summarization by ChatGPT in zero-shot settings without fine-tuning and language understanding of low resource language also remains a challenge with low evaluation results \cite{bang2023multitask}. Articles whose corresponding summaries (total 40; 20 each dataset) were evaluated by humans were input to ChatGPT with the prompt: "summarize it in Urdu in 1-3 sentences" along with \textit{article text}. Summaries generated by ChatGPT were evaluated by the same methodologies i.e. ROUGE, BERTScore, and Human. Fig. \ref{fig:ChatGPTComparison}) shows a comparison between abstractive summaries generated by our adapted urT5 model and ChatGPT model. The performance of ChatGPT (Dialogue based LM) on this task was lower than our model (designed specifically for summarization of a low-resource language Urdu). \footnote{Primary source of pre-training data used in GPT3 is CommonCrawl \url{https://commoncrawl.github.io/cc-crawl-statistics/plots/languages}; Training data of (low-resource) Urdu < 0.05\% as compared to (high-resource) English > 45\%}.

\section{Conclusion and Future Research}\label{sec:concl}
NLP has evolved significantly due to the recent inception of transformer-based architecture comprising of Deep Learning and Artificial Neural Networks. Automatic Summarization is a comparatively complex downstream task under the NLP umbrella due to various factors e.g. differences of opinion regarding the importance of information, absence of unanimous evaluation standards, etc. Moreover in Abstractive Summarization, new words are utilized which are not present in the vocabulary of the document to be summarized. This property of abstractive summarization presents endless possibilities for summarization making it difficult for automatic evaluation. Despite the challenges, the latest transformer-based models have proven their efficiency even in Automatic Summarization (both extractive and abstractive). Considering the lack of research in low-resource summarization a dataset has been created from publicly available sources which can be replicated for any low-resource language. Utilizing the newly created dataset and available multilingual models, a methodology was adopted for the adaptation of multilingual models for monolingual purposes efficiently with comparative evaluation results in a low-resource development setup that is freely available. Research has also been made available online which can be utilized for future experiments. A few of the future areas of research which demands exploration are:-
\begin{itemize}
\item \textit{Datasets} - Creation of quality datasets for training and evaluations including multi-domain datasets (comprising of a variety of sources, news, reviews, books, lectures, etc) and cross-lingual parallel datasets (same text in datasets in multiple languages) specifically to tackle problems of low resource languages.
\item \textit{Models} - Multilingual models (generic) which are capable of tackling the problem of under-representation of low-resource languages and able to understand more complex lingua.
\item \textit{Modular Approach} - Most of the available multilingual models are generally trained over multiple tasks of NLU and NLG with a large number of parameters resulting in larger sizes with more memory consumption. A modular approach towards models may be explored where the model while retaining the generalization of NLU and NLG tasks may be able to utilize modules/layers necessary for specific downstream tasks resulting in lesser resource utilization. One such technique of loading only monolingual vocabulary is used in this research however models inherently don't provide such flexibility.
\item \textit{Evaluation} - Available Evaluation methods for summarization are lacking research as compared to NLU and NLG tasks. These methods have inherent issues which are already quite frequently being discussed for high-resource languages. Authenticity \& verification of these evaluation methods for low-resource languages and their global applicability for cross-lingual purposes is altogether another avenue of research that still lacks progress.
\end{itemize}

\paragraph{Data Availability}
The datasets generated during and analyzed during the current study are available from the corresponding author on reasonable request.

\paragraph{Declaration}
\textbf{Conflict of Interest}: The authors report no conflicts of interest.

\bibliographystyle{unsrtnat}
\bibliography{summarization}  






\end{document}